\documentclass[pdflatex,sn-basic,Numbered]{sn-jnl}

\usepackage{amsmath}
\usepackage{amssymb}
\usepackage{amsfonts}
\usepackage{booktabs}
\usepackage{siunitx}
\usepackage{multirow} % 如果你的文档中还未包含 multirow
\usepackage{algorithm}
\usepackage{algorithmicx}
\usepackage{graphicx}
\usepackage{dblfloatfix}
\usepackage{array}
\usepackage{makecell}
\usepackage{longtable}
\usepackage{enumitem}
\usepackage{pdflscape}
\usepackage{adjustbox}
\usepackage{placeins}
\usepackage{algpseudocode}

\begin{document}

% Float placement tuning (reduces float-only pages and large blank areas)
\renewcommand{\topfraction}{0.9}
\renewcommand{\bottomfraction}{0.9}
\renewcommand{\textfraction}{0.07}
\renewcommand{\floatpagefraction}{0.7}
\renewcommand{\dblfloatpagefraction}{0.7}
\renewcommand{\dbltopfraction}{0.9}
\setcounter{topnumber}{2}
\setcounter{bottomnumber}{2}
\setcounter{totalnumber}{4}
\setcounter{dbltopnumber}{2}

% Main title of the paper
\title[Cluster-Aware Attention-Based DRL for PDP]{Cluster-Aware Attention-Based Deep Reinforcement Learning for Pickup and Delivery Problems}

%% Authors (Wentao Wang and Lifeng Han contributed equally.)
\author[1]{\fnm{Wentao} \sur{Wang}}\email{shiyanxi1@mail.dlut.edu.cn}
\equalcont{These authors contributed equally to this work.}

\author[1]{\fnm{Lifeng} \sur{Han}}\email{2369787465@mail.dlut.edu.cn}
\equalcont{These authors contributed equally to this work.}

\author*[2]{\fnm{Guangyu} \sur{Zou}}\email{zgy\_neu@hotmail.com}

\affil[1]{\orgdiv{Leicester International Institute}, \orgname{Dalian University of Technology}, \orgaddress{\city{Panjin}, \postcode{124221}, \state{Liaoning}, \country{China}}}

\affil[2]{\orgdiv{Department of Electronic and Information Technology}, \orgname{Dalian University of Technology}, \orgaddress{\city{Panjin}, \postcode{124221}, \state{Liaoning}, \country{China}}}

\abstract{The Pickup and Delivery Problem (PDP) is a fundamental and challenging variant of the Vehicle Routing Problem, characterized by tightly coupled pickup--delivery pairs, precedence constraints, and spatial layouts that often exhibit clustering. Existing deep reinforcement learning (DRL) approaches either model all nodes on a flat graph, relying on implicit learning to enforce constraints, or achieve strong performance through inference-time collaborative search at the cost of substantial latency. In this paper, we propose \emph{CAADRL} (Cluster-Aware Attention-based Deep Reinforcement Learning), a DRL framework that explicitly exploits the multi-scale structure of PDP instances via cluster-aware encoding and hierarchical decoding. The encoder builds on a Transformer and combines global self-attention with intra-cluster attention over depot, pickup, and delivery nodes, producing embeddings that are both globally informative and locally role-aware. Based on these embeddings, we introduce a Dynamic Dual-Decoder with a learnable gate that balances intra-cluster routing and inter-cluster transitions at each step. The policy is trained end-to-end with a POMO-style policy gradient scheme using multiple symmetric rollouts per instance. Experiments on synthetic clustered and uniform PDP benchmarks show that CAADRL matches or improves upon strong state-of-the-art baselines on clustered instances and remains highly competitive on uniform instances, particularly as problem size increases. Crucially, our method achieves these results with substantially lower inference time than neural collaborative-search baselines, suggesting that explicitly modeling cluster structure provides an effective and efficient inductive bias for neural PDP solvers.}

\maketitle

\section{Introduction}\label{sec:introduction}

The vehicle routing problem (VRP) is a fundamental combinatorial optimization problem with pervasive applications in modern transportation and logistics systems, including city-scale express delivery, ride-hailing and ride-sharing, e-commerce fulfillment, and on-demand food delivery~\cite{mourad2019survey,Parragh2008}. Emerging deployments further couple routing with partial information and operational constraints, as in online dispatching, crowdsourced logistics, and electrified fleets~\cite{yu2019online,arslan2019crowdsourced,lin2022evrptw}. Among its many variants, the Pickup and Delivery Problem (PDP) plays a central role because each transportation request is represented by a pickup--delivery pair that must be served by the same vehicle under strict precedence constraints~\cite{Parragh2008}; the resulting pairing and bipartite structure substantially complicate both modeling and algorithm design.

In this paper, we focus on a classical static single-vehicle PDP with a single depot and $n$ pickup--delivery pairs defined in a 2D Euclidean space. The objective is to construct a Hamiltonian tour that starts and ends at the depot, visits every pickup and delivery node exactly once, and respects all precedence constraints, while minimizing the total travel distance. This setting coincides with pickup-and-delivery traveling salesman variants commonly used as testbeds in the PDP literature and provides a clean environment for isolating the effect of policy architectures.

Over the past decades, a rich body of exact and heuristic methods has been developed for PDP and related VRPs~\cite{Parragh2008,ghilas2016alns}. While exact approaches provide optimal or near-optimal benchmarks, they scale poorly as the number of requests and side constraints grows, and high-performing heuristics often rely on hand-crafted neighborhoods and careful parameter tuning.

Motivated by the desire to automate heuristic design, deep reinforcement learning (DRL) and neural combinatorial optimization have also shown strong results for routing. Attention-based encoder--decoder construction policies and POMO-style training achieve near state-of-the-art performance on the TSP and capacitated VRP (CVRP)~\cite{vinyals2015pointer,bello2016nco,kool2019attention,nazari2018rl4vrp,kwon2020pomo}, and search-augmented/improvement paradigms further boost solution quality~\cite{chen2019neurewriter,wu2022improvement,sun2024difusco}. For PDP, incorporating pickup--delivery pairing and precedence remains nontrivial, and recent methods introduce role-aware attention or learned neighborhood search (e.g., Heter-AM, ENNS, NCS)~\cite{li2022heter,ma2021enns,kong2024ncs}; Section~\ref{sec:related} provides a detailed review.

Despite these advances, existing PDP neural solvers still struggle to exploit multi-scale spatial regularities commonly observed in practice. Most architectures operate on a flat node graph and must infer region-level structure implicitly, even though pickup and delivery locations often form coherent zones where good routes alternate between intra-zone exploitation and inter-zone transitions. Moreover, a single decoder is typically used to handle both local and global routing decisions, while search-based frameworks can improve solution quality at the cost of multiple inference-time iterations and increased latency.

These observations motivate the approach proposed in this paper. We introduce \emph{CAADRL}, a cluster-aware DRL framework for the single-vehicle PDP, based on the hypothesis that explicitly modeling the multi-scale structure between depot, pickup, and delivery regions provides an effective inductive bias. CAADRL combines a cluster-aware Transformer encoder (global self-attention plus intra-cluster attention under a cluster mask) with a hierarchical Dynamic Dual-Decoder coordinated by a learnable gate, enabling one-pass autoregressive construction without iterative improvement; the policy is trained end-to-end under a POMO-style regime adapted to PDP.

The main contributions of this work are summarized as follows:
\begin{itemize}
  \item \textbf{Cluster-aware encoder architecture:} We propose a cluster-aware Transformer encoder with a \emph{Cluster-Aware Attention} mechanism that fuses global self-attention with intra-cluster attention restricted by a cluster mask at every layer. The resulting embeddings are simultaneously globally consistent and locally role-aware, making explicit the natural separation between depot, pickup, and delivery regions and providing a multi-scale representation of PDP instances in Euclidean space.
  \item \textbf{Hierarchical decoding with dual decoders and gating:} We design a \emph{Dynamic Dual-Decoder} framework in which one decoder specializes in intra-cluster routing decisions and the other in inter-cluster transitions. A cluster-level gating network produces a soft decision at each step on whether to stay within the current cluster or switch to another cluster, leading to globally consistent routes that respect precedence constraints while effectively exploiting region-level structure. Importantly, our method remains a pure construction policy, requiring only a single autoregressive decoding pass per solution without iterative neural improvement loops.
  \item \textbf{POMO-based training and comprehensive evaluation:} We extend the POMO training paradigm~\cite{kwon2020pomo} to cluster-aware PDP policies, enabling stable and sample-efficient learning by generating multiple diverse rollouts per instance in a single forward pass. Extensive experiments on synthetic clustered and uniform PDP datasets with varying sizes (PDP10, PDP20, PDP40, and PDP80), as well as comparisons with a heterogeneous-attention baseline~\cite{li2022heter} and NCS~\cite{kong2024ncs}, show that the proposed method matches or improves upon strong state-of-the-art baselines on clustered instances and remains highly competitive on uniform instances, with favorable inference times, especially at larger problem scales.
\end{itemize}

The remainder of this paper is organized as follows. Section~\ref{sec:related} reviews exact, heuristic, and learning-based methods for PDP and related VRPs. Section~\ref{sec:preliminary} introduces the formal problem definition and mixed-integer formulation. Section~\ref{sec:methodology} presents the proposed DRL methodology in detail, including the cluster-aware encoder, hierarchical dual-decoder, and POMO-based training scheme. Section~\ref{sec:experiments} describes the experimental setup, reports numerical results on both clustered and uniform PDP benchmarks, and analyzes scalability and generalization. Section~\ref{sec:conclusion} concludes the paper and outlines directions for future research.

\section{Related Work}\label{sec:related}

In this section, we review exact and heuristic methods for PDP and related VRPs, and then discuss recent learning-based approaches with a focus on DRL and neural combinatorial optimization.

\subsection{Exact and Heuristic Methods}

The PDP extends classical VRPs by coupling pickups and deliveries with pairing and precedence constraints. Recent surveys in transportation and OR summarize model variants, exact formulations, and application-driven extensions in shared mobility and on-demand logistics~\cite{Parragh2008,mourad2019survey}. Exact approaches based on mixed-integer programming and branch-and-price/branch-and-cut remain indispensable for small and medium instances and provide optimal benchmarks. Recent Transportation Science studies develop exact formulations for pickup-and-delivery traveling salesman variants with neighborhood structure, where each request corresponds to a region and geometric flexibility can be exploited to tighten formulations and prune search~\cite{gao2023exact}. Exponential-size neighborhood models further embed very large move sets into exact algorithms, leading to stronger bounds and more effective search without explicit enumeration of all neighborhood moves~\cite{pacheco2023expneigh}. Polyhedral advances for capacitated multi-pickup and delivery with time windows derive strong cutting planes that tighten LP relaxations and reduce branch-and-cut burden, underscoring the value of constraint-specific inequalities beyond basic precedence and degree constraints~\cite{kohar2023cutting}. Complementary Transportation Science studies show that branch-and-cut-and-price can also solve richer PDP variants such as multi-agent and multicompartment settings, but the computational burden still grows quickly with instance size and constraint complexity~\cite{lam2024bcp,aerts2024bpc}.

To scale to larger instances, heuristic frameworks dominate. Constructive insertion rules and local search moves are typically embedded within LNS/ALNS schemes that alternate destroy and repair operations over large neighborhoods; Ghilas \emph{et al.} demonstrate the effectiveness of ALNS for PDPs with time windows and scheduled lines~\cite{ghilas2016alns}. Recent LNS variants also handle split loads and transshipments, where servicing a request may be distributed across multiple visits or transfer points, increasing coupling across route segments and requiring coordinated destroy/repair operators to maintain feasibility~\cite{li2016adaptive}. Despite strong empirical performance, such heuristics depend on hand-crafted neighborhoods, scoring functions, and penalty tuning, and they often require substantial redesign when operational constraints shift, motivating hybridization with learning-based components~\cite{zhao2021hybrid}.

\subsection{Learning-Based Methods}

Learning-based routing has evolved from pointer-network policies to attention-based encoder--decoder models trained with policy gradients, enabling end-to-end construction of tours from node coordinates~\cite{vinyals2015pointer,bello2016nco,nazari2018rl4vrp,kool2019attention}. POMO further stabilizes training by exploiting permutation symmetries via multiple rollouts per instance~\cite{kwon2020pomo}. A broader survey of RL for combinatorial optimization situates these methods within a growing landscape of neural solvers and training strategies~\cite{mazyavkina2021survey}.

Beyond one-shot construction, neural improvement methods learn to refine solutions via local edits or learned neighborhoods. NeuRewriter and learning-based improvement heuristics introduce rewrite or move-selection policies that mimic classical local search~\cite{chen2019neurewriter,wu2022improvement}. Neural large neighborhood search and flexible neural $k$-opt integrate learned destroy--repair or $k$-opt operators into iterative optimization pipelines; latent search spaces via variational autoencoders offer another route to structure exploration~\cite{hottung2022nlns,ma2023neuopt,hottung2021vae}. Diffusion-based solvers learn to sample high-quality solutions from learned distributions~\cite{sun2024difusco}. Hybrid approaches that couple neural construction with classical local search also report consistent gains on routing benchmarks~\cite{zhao2021hybrid}.

For PDP, specialized models explicitly encode pickup--delivery structure. Heter-AM introduces role-specific attentions and precedence-aware masking to distinguish depot, pickup, and delivery nodes in the policy network~\cite{li2022heter}. ENNS learns PDP-tailored neighborhood operators and a learned selection policy to guide search~\cite{ma2021enns}. NCS further couples a construction model with a ruin-and-repair improvement module trained collaboratively, delivering state-of-the-art quality at the cost of multiple inference-time improvement iterations~\cite{kong2024ncs}. Recent studies also extend DRL to dynamic, flexible, and backhaul-rich pickup--delivery settings, underscoring the need for policies that generalize across operational constraints~\cite{xiang2024crowdshippers,tian2025flexible,wang2025backhauls}.

However, existing PDP neural solvers largely treat interactions on a flat graph and decode with a single decision module, so region-level structure is captured only implicitly. Search-based variants can mitigate this by iteratively improving a constructed tour, but they increase inference-time latency. These limitations motivate our cluster-aware encoder and hierarchical dual-decoder, which separate intra-cluster and inter-cluster decisions while maintaining one-pass decoding efficiency.

%, PRELIMINARY Section,
\section{Preliminary}\label{sec:preliminary}
The Pickup and Delivery Problem (PDP) is a classic variant of the Vehicle Routing Problem (VRP). Given a single depot and $n$ customer requests, each request consists of a pickup point and a delivery point that must be served by the same vehicle. We define the set of pickup points as $\mathbf{P} = \{x_i\}_{i=1}^{n}$ and the set of delivery points as $\mathbf{D} = \{x_{i+n}\}_{i=1}^{n}$. Here, nodes $x_i$ and $x_{i+n}$ form a logically coupled pair that must satisfy a precedence constraint, meaning that the pickup node $x_i$ must always be visited before its corresponding delivery node $x_{i+n}$. If we define the depot as node $x_0$, the complete set of nodes in the problem is
\[
    \mathbf{X} = \{x_0\} \cup \mathbf{P} \cup \mathbf{D},
\]
and a feasible solution is a permutation of all nodes in $\mathbf{X}$ that respects the precedence constraints.

The objective of this problem is to plan a single-vehicle route that starts from the depot $x_0$, visits each customer node exactly once, and finally returns to the depot, while minimizing the total travel distance (equivalently, the total travel time under a constant speed). This classical static PDP setting provides a clean testbed for evaluating our proposed deep reinforcement learning model and for isolating the contribution of the learned policy architecture.

\begin{figure*}[!t]
    \centering
    \includegraphics[width=\linewidth]{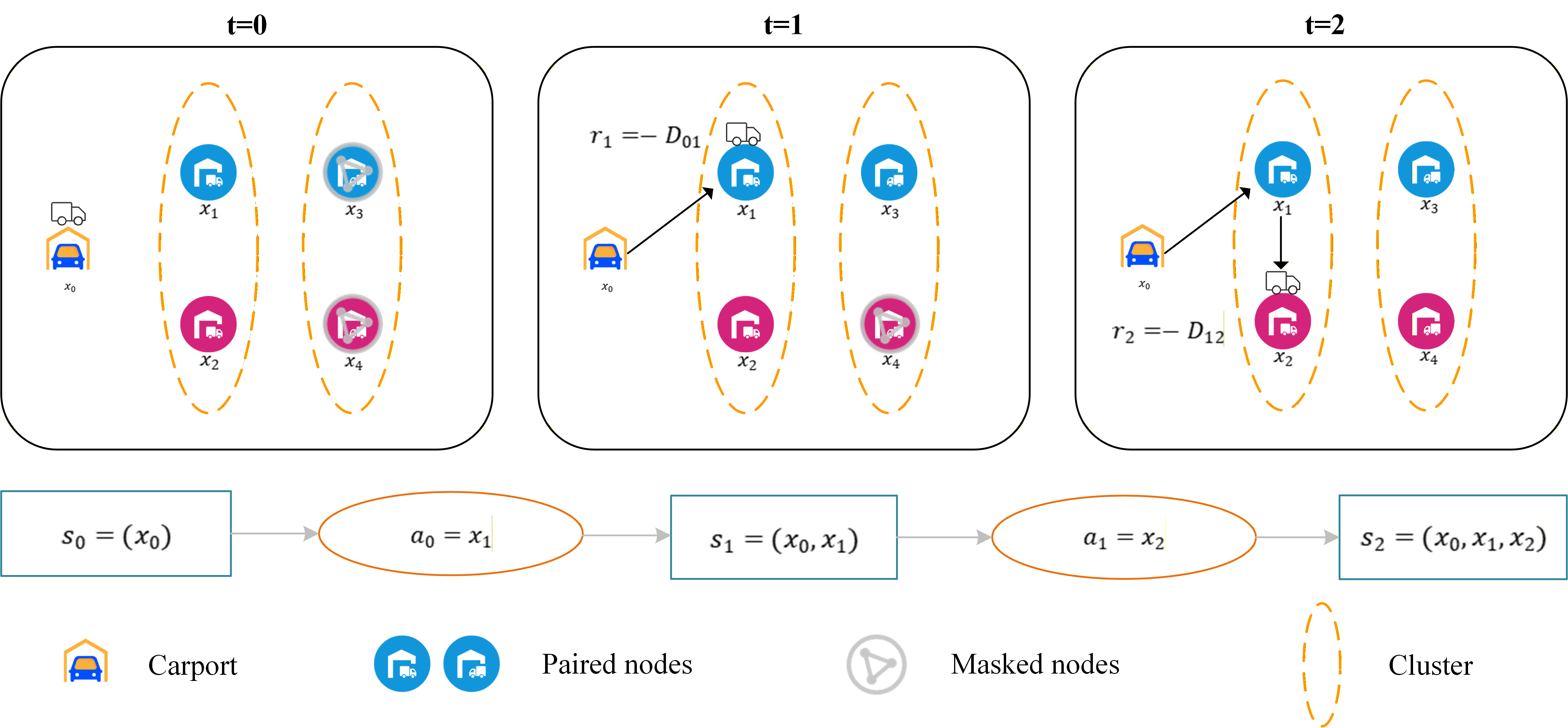} % 请确保文件名正确
    \caption{Illustration of the sequential decision-making process for PDP modeled as an MDP. The process evolves from $t=0$ to $t=2$, showing the transition of states $s_t$ based on selected actions $A_t$ (visiting nodes). The dashed orange ovals highlight the spatial cluster structure (pickup and delivery regions) explicitly considered in our framework. The reward $r_{t+1}$ corresponds to the negative distance traveled.}
    \label{fig:mdp_process}
\end{figure*}

Let $D_{ij}$ be the Euclidean distance between nodes $x_i$ and $x_j$, and let $f$ denote the (constant) vehicle speed. Let the binary decision variable be $y_{ij} \in \{0, 1\}$, which is 1 if and only if the vehicle travels directly from node $i$ to node $j$ in the final route. Furthermore, let $B_i$ denote the time of arrival at node $i$ (or, equivalently, the cumulative travel time up to node $i$). The PDP considered in this paper can be formulated as the following mixed-integer program:
\begin{equation}
    \min \sum_{i \in \mathbf{X}} \sum_{j \in \mathbf{X}} \frac{D_{ij}}{f} y_{ij}
    \label{eq:objective}
\end{equation}

subject to the following constraints:
\begin{align}
    \sum_{j \in \mathbf{X}, j \neq i} y_{ij} &= 1, && \forall i \in \mathbf{X} \label{eq:const_out_degree} \\
    \sum_{i \in \mathbf{X}, i \neq j} y_{ij} &= 1, && \forall j \in \mathbf{X} \label{eq:const_in_degree} \\
    B_j &\ge B_i + \frac{D_{ij}}{f} - M(1 - y_{ij}), && \forall i, j \in \mathbf{X} \label{eq:const_time_window} \\
    B_{i+n} &\ge B_i, && \forall x_i \in \mathbf{P} \label{eq:const_precedence}
\end{align}
Here, constraints \eqref{eq:const_out_degree} and \eqref{eq:const_in_degree} enforce degree balance at each node and ensure that each node is visited exactly once in the resulting route. Constraint \eqref{eq:const_time_window} defines the arrival time relationship between two consecutively visited nodes and plays the role of a sequencing constraint: if edge $(i,j)$ is used (i.e., $y_{ij} = 1$), then the arrival time at node $j$ must be at least the arrival time at node $i$ plus the travel time $\tfrac{D_{ij}}{f}$; otherwise, the constraint is relaxed by the large constant $M$. This big-$M$ relaxation can be viewed as an auxiliary solver-side penalty/relaxation term commonly introduced in exact MIP formulations to eliminate subtours; in our learning-based solver we do not optimize the MIP directly and instead enforce feasibility during autoregressive decoding via masking. Constraint \eqref{eq:const_precedence} is the core precedence constraint that ensures the pickup node $x_i$ is always visited before the corresponding delivery node $x_{i+n}$. Together, these constraints define the feasible solution space over which our learning-based method will operate.

%, METHODOLOGY Section,

\section{Methodology}\label{sec:methodology}
In this section, we first reformulate the PDP as a Reinforcement Learning (RL) problem so that route construction can be treated as a sequential decision-making process. We then detail the encoder--decoder based policy network designed to solve it. The core innovation of this network is a novel architecture that explicitly models and leverages the inherent ``cluster'' structure of the PDP. This is achieved through two main components that work in a tightly coupled manner. First, in the encoder, we design a Cluster-Aware Attention mechanism that generates node embeddings with both a global perspective over the entire instance and fine-grained role-awareness with respect to depot, pickup, and delivery clusters. Second, in the decoder, these powerful embeddings are consumed by a Hierarchical Decision Framework. This framework, realized through a Dynamic Dual-Decoder architecture, handles tactical intra-cluster movements and strategic inter-cluster transitions separately, with their outputs coordinated by a learnable gating mechanism. Finally, the whole network is trained end-to-end under the Policy Optimization with Multiple Optima (POMO) framework, which provides a sample-efficient and stable training procedure.

\subsection{Reformulation as RL Form}
To solve the PDP, the route construction process can be viewed as a sequential decision-making process, making it natural to formalize it as an RL problem~\cite{bello2016nco,nazari2018rl4vrp,kool2019attention}. Fig.~\ref{fig:mdp_process} illustrates this process, visualizing how the partial tour evolves over time steps $t=0, 1, 2$ within the underlying cluster structure. Specifically, we model this process as a Markov Decision Process (MDP), with its core components defined as follows:

\textbf{State.} The state $s_t = (\tau_{0:t})$ at decision step $t$ represents the partial solution constructed so far, where $\tau_{0:t}$ is the ordered sequence of visited nodes up to the current step, and $\tau_0$ denotes the depot. This state compactly encodes both the current vehicle location and the set of nodes that have already been visited (implicitly capturing which nodes remain feasible).

\textbf{Action.} Let $\mathcal{A}(s_t)$ denote the feasible action set at step $t$. An action $A_t \in \mathcal{A}(s_t)$ corresponds to selecting the next node to visit, i.e., $\tau_{t+1}$, from the set of all currently unvisited and constraint-satisfying nodes. The feasibility of an action is jointly determined by whether the node has been visited, whether the precedence constraint between each pickup and its paired delivery is satisfied, and whether the route construction has not yet terminated.

\textbf{Transition.} When action $A_t = \tau_{t+1}$ is executed in state $s_t$, the system transitions deterministically to a new state $s_{t+1} = (\tau_{0:t+1})$. This transition simply appends the newly selected node to the partial tour and updates the set of feasible nodes for the next decision step.

\textbf{Reward.} To minimize the total travel distance, we define the reward as the negative of the total path length. The reward is received after each step. Specifically, suppose at decision step $t$, the vehicle is at node $\tau_t$ and executes action $A_t$ to travel to the next node $\tau_{t+1}$. The incremental reward $r_{t+1}$ received after this transition is defined as:
    \begin{equation}
        r_{t+1} = r(s_t, A_t) = -D_{\tau_t, \tau_{t+1}}.
    \end{equation}
    The total reward $R$ is the sum of all incremental rewards over the entire closed tour, including the return to the depot, and is calculated as $R = \sum_{t=0}^{|\mathbf{X}|-1} r_{t+1}$. Maximizing this cumulative reward, is therefore equivalent to minimizing the total route length.

\textbf{Policy.} Our goal is to learn a stochastic policy $\pi_{\theta}(A_t|s_t)$, parameterized by $\theta$, which automatically selects at each timestep a node that satisfies all problem constraints. Based on the chain rule, the probability of a complete solution (i.e., a full permutation of nodes $\tau$) can be decomposed as:
    \begin{equation}
        P(\tau|\mathbf{X}; \theta) = \prod_{t=0}^{|\mathbf{X}|-1} \pi_{\theta}(\tau_{t+1}|\mathbf{X}, \tau_{0:t}).
    \end{equation}

In our framework, this policy is represented by a differentiable encoder--decoder architecture, which allows us to optimize $\theta$ using gradient-based RL algorithms.

\begin{figure}[tbp]
    \centering
    \includegraphics[width=\columnwidth]{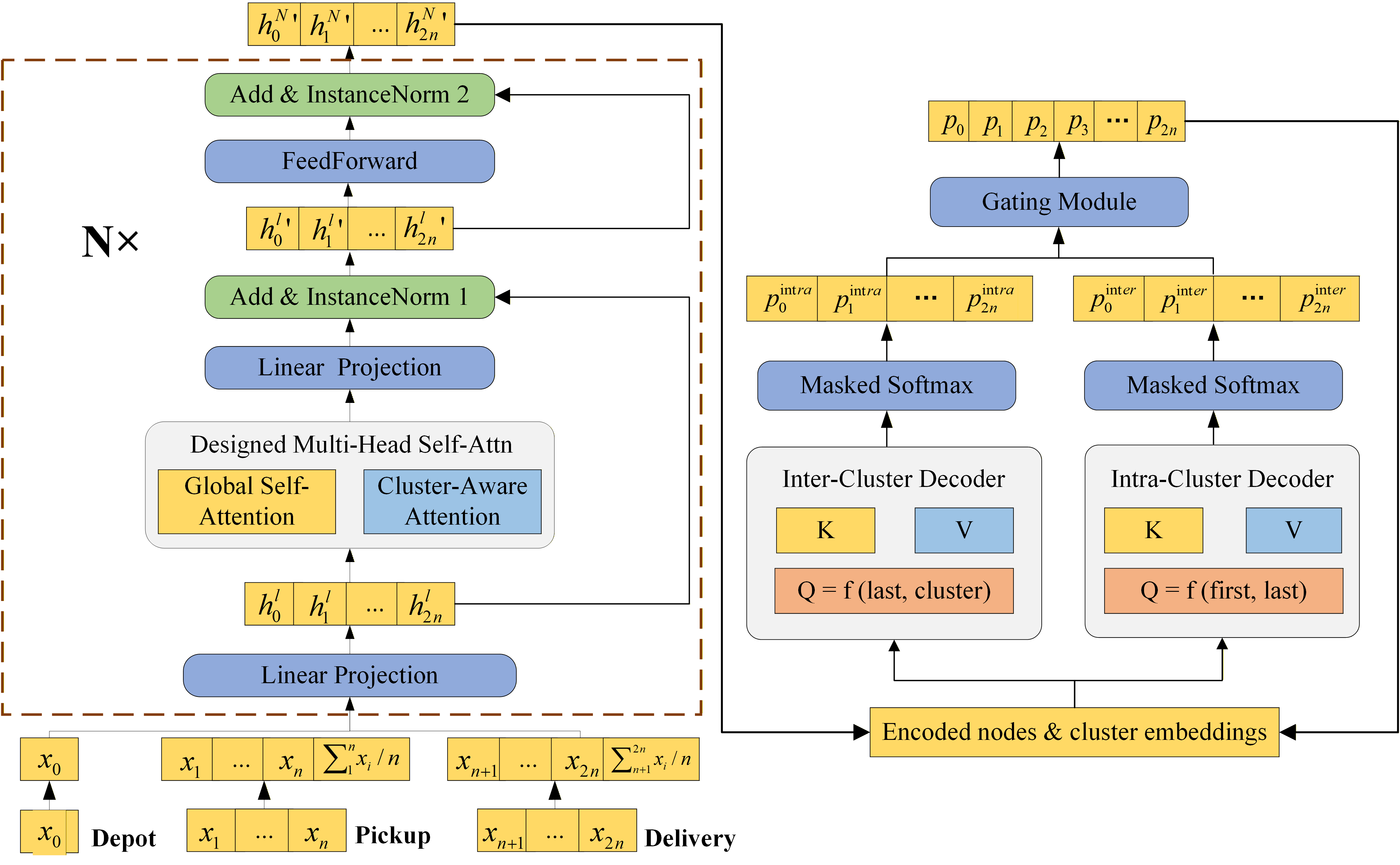}
    \caption{Cluster-Aware Policy Network architecture.}
    \label{fig:architecture}
\end{figure}

\subsection{Policy Network Architecture}
We design a Transformer-based encoder--decoder model to parameterize the policy $\pi_{\theta}$~\cite{kool2019attention}.
The overall architecture of our proposed model is illustrated in Fig.~\ref{fig:architecture}. 

The core of this model lies in its deep understanding and utilization of the intrinsic ``cluster'' structure of the PDP (i.e., the pickup cluster and the delivery cluster, together with the depot). Concretely, the encoder produces cluster-aware node embeddings, and the decoder performs hierarchical decision making based on these embeddings.

\subsubsection{Encoder with Cluster-Aware Attention}
The encoder's role is to generate context-rich embedding representations $h_i$ for all input nodes that capture both global spatial relationships and local role-specific patterns. It consists of an initial embedding layer, a cross-attention layer centered at the depot, and a stack of $L$ main encoder layers, each containing two parallel multi-head attention branches.

The initial features of each node $x_i$ are composed of its 2D coordinates and its discrete cluster ID (0 for depot, 1 for pickup, 2 for delivery). These raw features are first mapped to a high-dimensional embedding space via a learnable linear layer, yielding an initial embedding for every node in $\mathbf{X}$.

Before entering the main encoder, we first enhance the depot's representation using a cross-attention layer. In this layer, the depot node's embedding serves as the Query, while the embeddings of all customer nodes (both pickup and delivery) serve as the Keys and Values. This process generates a more informative depot representation that incorporates global information about the customer cluster, enabling the model to better capture the overall geometry of the instance and the relative layout of pickups and deliveries.

Subsequently, the embeddings of all nodes (including the updated depot node) are fed into a multi-layer Transformer encoder. Unlike a standard Transformer encoder, each of our layers performs two functionally specialized multi-head attention computations in parallel to simultaneously capture both the global structure of the problem and local intra-cluster relationships. Let $H^{(l-1)} \in \mathbb{R}^{(|\mathbf{X}|) \times d_h}$ be the node embedding matrix from the previous layer. In layer $l$, the following are computed:

\textbf{Global Self-Attention.} This mechanism is identical to the standard self-attention in a Transformer. It allows each node to attend to all other nodes in the input sequence without any restrictions. Its purpose is to learn global spatial dependencies and the overall path structure across the entire graph. Its Query, Key, and Value matrices are generated by a set of independent, learnable weight matrices $(W_Q^G, W_K^G, W_V^G)$:
\begin{equation}
    Q_G = H^{(l-1)}W_Q^G, \quad K_G = H^{(l-1)}W_K^G, \quad V_G = H^{(l-1)}W_V^G.
\end{equation}
The output of this part, $H_G$, is calculated by the multi-head attention function, which captures dependencies by performing scaled dot-product attention across $h$ heads:
\begin{equation}
\begin{split}
    H_G &= \text{MultiHead}(Q_G, K_G, V_G) \\
        &= \text{Concat}(\text{head}_1^G, \dots, \text{head}_h^G)W^O_G,
\end{split}
\end{equation}
where each head is computed as:
\begin{equation}
    \text{head}_i^G = \text{softmax}\left(\frac{Q_G^{(i)}(K_G^{(i)})^T}{\sqrt{d_k}}\right)V_G^{(i)}.
\end{equation}

\textbf{Intra-Cluster Attention.} To enable the model to explicitly learn specific relationships among nodes of the same type (e.g., among all pickup points or among all delivery points), we introduce intra-cluster attention. This mechanism uses a pre-computed structural mask $M^{\text{cluster}}$ to restrict each node to attend only to nodes within its own cluster. This allows the model to learn refined, role-specific representations that highlight local structures such as the internal geometry of the pickup region or the delivery region. This attention mechanism uses another set of independent weight matrices $(W_Q^C, W_K^C, W_V^C)$:
\begin{equation}
    Q_C = H^{(l-1)}W_Q^C, \quad K_C = H^{(l-1)}W_K^C, \quad V_C = H^{(l-1)}W_V^C.
\end{equation}
Its attention scores $a_{ij}^C$ incorporate the cluster mask before the Softmax function is applied to ensure role-specific focus. Specifically, the output $H_C$ is defined as:
\begin{equation}
\begin{split}
    H_C &= \text{MultiHead}(Q_C, K_C, V_C, M^{\text{cluster}}) \\
        &= \text{Concat}(\text{head}_1^C, \dots, \text{head}_h^C)W^O_C,
\end{split}
\end{equation}
where the attention for each head $i$ is computed with the cluster mask as:
\begin{equation}
    \text{head}_i^C = \text{softmax} \left( \frac{Q_C^{(i)} (K_C^{(i)})^T}{\sqrt{d_k}} + M^{\text{cluster}} \right) V_C^{(i)}.
\end{equation}
The mask is defined such that $M^{\text{cluster}}_{ij} = 0$ if nodes $i$ and $j$ belong to the same cluster, and $-\infty$ otherwise, effectively zeroing out the attention weights for nodes in different clusters. Note that this is a structural mask (cluster membership), not a causal mask.

The outputs of these two parallel attention mechanisms, $H_G$ and $H_C$, are then added element-wise to fuse global and local contextual information:
\begin{equation}
    H_{\text{fused}} = H_G + H_C.
\end{equation}
The fused representation $H_{\text{fused}}$ passes through a final linear combination layer, followed by a residual connection and layer normalization, and is then fed into a position-wise feed-forward neural network to produce the output embedding $H^{(l)}$ for that encoder layer. Stacking $L$ such layers yields the final node embeddings $h_i$ used by the decoder. This dual-attention design endows the embeddings with both a macroscopic global perspective and microscopic role-awareness, which is crucial for reasoning about cluster-dependent routing patterns.

\subsubsection{Decoder}
Given the node embedding matrix $H \in \mathbb{R}^{|\mathbf{X}| \times d_h}$ from the encoder, the decoder's goal is to sequentially generate a probability vector $p_t$ at each decision step $t$ for selecting from the currently feasible nodes. Our model achieves this through a three-stage process involving hierarchical query generation, parallel decoding pipelines, and learned gating fusion, thereby explicitly separating intra-cluster exploitation from inter-cluster exploration.

At the beginning of each decoding step, the model dynamically constructs two different query vectors that reflect the hierarchical nature of the decision:

\textbf{Intra-Cluster Query ($q_{\text{intra}}$): }When making tactical intra-cluster decisions, the query vector is defined by the embeddings of the start node $h_{\text{first}}$ and the end node $h_{\text{last}}$ of the current sub-path, so as to capture local path directionality and the current progression within the cluster:
    \begin{equation}
        q_{\text{intra}} = W_q^{\text{first}}h_{\text{first}} + W_q^{\text{last}}h_{\text{last}}.
    \end{equation}

\textbf{Inter-Cluster Query ($q_{\text{inter}}$):} When making strategic inter-cluster decisions, the query vector is defined by the embedding of the last node $h_{\text{last}}$ and the mean embedding of its cluster $\bar{h}_{\text{cluster}}$. This combination allows the model to merge macroscopic cluster-level information with the current local context:
    \begin{equation}
        q_{\text{inter}} = W_q^{\text{last}}h_{\text{last}} + W_q^{\text{cluster}}\bar{h}_{\text{cluster}}.
    \end{equation}

Here, all $W_q$ are learnable parameter matrices. In this way, the model explicitly constructs two queries that emphasize different decision scales.

The two generated query vectors, $q_{\text{intra}}$ and $q_{\text{inter}}$, are then fed into two independent but parameter-sharing decoding pipelines, each generating a complete probability distribution over the feasible nodes. For any given query $q$, the decoding pipeline proceeds as follows:

First, a multi-head attention mechanism calculates an attention output vector $h^g$. The core of this process is to compute compatibility scores between the query $q$ and the key matrix $K=HW_K$ of all nodes, and then use these scores to weight the value matrix $V=HW_V$. The attention weight vector $a$ is calculated as:
    \begin{equation}
        a = \text{softmax}\left(\frac{q K^T}{\sqrt{d_k}} + M_{\text{mask}}\right),
    \end{equation}
    where $M_{\text{mask}}$ is used to mask out illegal nodes that violate feasibility constraints. The attention output $h^g$ is then obtained by a weighted sum over the value matrix $V$:
    \begin{equation}
        h^g = aV.
    \end{equation}
    
Next, the attention output vector $h^g$ is projected into a logits space with the same dimension as the number of nodes. This step is computed using a scaled dot-product form similar to that within the attention mechanism:
    \begin{equation}
        (u)_i = \frac{(h^g)^T k_i}{\sqrt{d_k}},
    \end{equation}
    where $k_i$ is the key vector of node $i$.

To enhance exploration and stabilize training, we clip the logits vector $u$ with a parameter $C$, yielding $\hat{u} = C \cdot \tanh(u)$. Then, the constraint mask $M_{\text{mask}}$ is applied once again, and the final result is converted into a valid probability distribution $P$ via the Softmax function:
\[
    P = \text{softmax}(\hat{u} + M_{\text{mask}}).
\]

Through the pipeline described above, we obtain two probability distributions in parallel: $P_{\text{intra}} = \text{Pipeline}(q_{\text{intra}})$ and $P_{\text{inter}} = \text{Pipeline}(q_{\text{inter}})$, each reflecting a different routing intention.

To dynamically fuse these two decision intentions, we introduce a learnable gating module $G$. This module takes necessary strategic information (such as $h_{\text{last}}, \bar{h}_{\text{current\_cluster}}, \bar{h}_{\text{other\_cluster}}$) as input and outputs a probability of ``staying'' in the current cluster, $p_{\text{stay}}$. Intuitively, $p_{\text{stay}}$ serves as a soft switch between intra-cluster exploitation and inter-cluster exploration.

The final action distribution $\pi_{\theta}(\cdot|\mathbf{X}, \tau_{0:t-1})$ is then expressed as a convex combination of the probability distributions output by the two decoders:
\begin{equation}
    \pi_{\theta}(\cdot|\mathbf{X}, \tau_{0:t-1}) = p_{\text{stay}} \cdot P_{\text{intra}} + (1 - p_{\text{stay}}) \cdot P_{\text{inter}},
\end{equation}
where the $i$-th element $\pi_i^t$ of the resulting distribution represents the probability of selecting node $x_i$ at decision step $t$. This hierarchical decision process is iterated until all nodes have been visited and the route is completed.

Regarding the decoding strategy at inference time, we can either adopt a greedy approach by selecting the node with the highest probability at each step, or generate multiple solutions by sampling from the probability distribution and then returning the best one. We will investigate and compare these strategies empirically in the experiments section.

\subsection{Training with POMO}
We employ the REINFORCE algorithm with a baseline to train the policy network and deeply integrate the POMO framework~\cite{kwon2020pomo} to enhance training stability and efficiency. The pseudocode for this training method is summarized in Algorithm~\ref{alg:pomo_rl_professional}.

The core idea of POMO is to effectively reduce the variance of the policy gradient by generating multiple different solutions for the same problem instance in a single forward pass. For a given problem instance $X^b$, we run $N$ decoding processes in parallel, each starting from a different customer node, to generate $N$ different solutions $\{\tau^{b,j}\}_{j=1}^N$ and their corresponding total rewards $\{R^{b,j}\}_{j=1}^N$. These parallel rollouts exploit the inherent symmetry of the routing problem and provide richer learning signals without additional forward passes.

Instead of using a separate critic network, we adopt a more direct and efficient shared baseline $b(X^b)$, which is defined as the empirical mean of the rewards obtained from these $N$ rollouts:
\begin{equation}
    b(X^b) = \frac{1}{N} \sum_{j=1}^{N} R^{b,j}.
\end{equation}
Therefore, the advantage function $A(\tau^{b,j}, X^b)$ for the $j$-th trajectory is the difference between its reward and the shared baseline: $R^{b,j} - b(X^b)$. The model parameters $\theta$ are updated using the Adam optimizer, and the gradient of the objective function on a batch can be approximated as:
\begin{equation}
    \nabla_{\theta} J(\theta) \approx \frac{1}{B} \sum_{b=1}^{B} \frac{1}{N} \sum_{j=1}^{N} (R^{b,j} - b(X^b)) \nabla_{\theta} \log \pi_{\theta}(\tau^{b,j}|X^b),
\end{equation}
where $B$ is the batch size. This method leverages multiple samples generated from a single model evaluation, which greatly stabilizes the training process and accelerates policy convergence while keeping the overall computational cost manageable.

\begin{algorithm}[!htbp]
\caption{POMO-based Reinforcement Learning Algorithm}
\label{alg:pomo_rl_professional}
\begin{algorithmic}[1]
    \State \textbf{Input:} number of epochs $I$; batch size $B$; number of parallel rollouts $N$
    \State \textbf{Input:} policy network $\pi_{\theta}$ with parameters $\theta$; Adam optimizer with learning rate $\eta$

    \For{epoch = 1 to $I$}
        \State Generate a batch of $B$ problem instances $\{X^b\}_{b=1}^B$ randomly
        
        \For{each instance $X^b$ in the batch}
            \State \textit{// POMO: Generate N parallel solutions by sampling from the policy}
            \For{$j = 1$ to $N$}
                \State Generate trajectory $\tau^{b,j}$ and its total reward $R^{b,j}$ by sampling from 
                \Statex \qquad $\pi_{\theta}(\cdot|X^b)$, each starting from a unique customer node.
            \EndFor
            
            \State \textit{// Calculate the shared baseline from the N rollouts}
            \State $b(X^b) \leftarrow \frac{1}{N} \sum_{j=1}^{N} R^{b,j}$
            
            \State \textit{// Calculate the loss for the current instance}
            \State $\mathcal{L}^b(\theta) \leftarrow -\frac{1}{N} \sum_{j=1}^{N} (R^{b,j} - b(X^b)) \log \pi_{\theta}(\tau^{b,j}|X^b)$
        \EndFor
        
        \State \textit{// Update the policy network using the averaged batch loss}
        \State $\mathcal{L}_{\text{batch}}(\theta) \leftarrow \frac{1}{B} \sum_{b=1}^{B} \mathcal{L}^b(\theta)$
        \State Update $\theta$ using Adam: $\theta \leftarrow \text{Adam}(\theta, \nabla_{\theta}\mathcal{L}_{\text{batch}}(\theta), \eta)$
    \EndFor
\end{algorithmic}
\end{algorithm}

\section{Computational Experimentation and Analysis}\label{sec:experiments}

In this section, we conduct a series of extensive computational experiments to comprehensively evaluate the performance of our proposed deep reinforcement learning-based model for solving the PDP. Our experiments are designed to answer the following key questions: 1) Is our hierarchical decoding framework effective in exploiting the cluster structure of PDP instances? 2) How does our model's performance compare to existing state-of-the-art deep learning methods on standard PDP benchmarks? 3) What is the generalization capability of our model across different data distributions, especially when the explicit cluster structure disappears and the instances resemble generic random PDPs?

\subsection{Experimentation Settings}\label{subsec:exp_settings}

Following common practices in existing research~\cite{kool2019attention,kwon2020pomo}, all problem instances are synthetically generated on a 2D plane $[0, 1] \times [0, 1]$. Distances between nodes are calculated using the Euclidean distance. We consider two different data distributions to jointly evaluate the \emph{specialization} and \emph{robustness} of our model:

\textbf{Clustered Distribution.} This is our primary testing scenario and is specifically aligned with the inductive bias of our model. All pickup points are independently sampled from a normal distribution centered at $(0.25, 0.25)$, while all delivery points are independently sampled from another normal distribution centered at $(0.75, 0.75)$. The standard deviation for both distributions is set to $0.1$. This construction yields two clearly separated spatial clusters, mimicking real-world situations where pickups and deliveries are geographically concentrated in different regions (for example, residential districts versus central business areas). By enforcing such a structured layout, this distribution highlights the need for coordinated intra-cluster routing and inter-cluster transitions and thus provides a stringent test of whether our cluster-aware encoder and hierarchical decoder can effectively exploit this structure.

\textbf{Uniform Distribution.} For comparison, we also generate datasets where all nodes (including the depot and all pickup/delivery nodes) are uniformly distributed within the $[0, 1] \times [0, 1]$ plane. In this setting there is no obvious cluster structure; pickup and delivery nodes are spatially mixed. This distribution serves two purposes: first, it evaluates the model's generalization performance on more random, unstructured scenarios; second, it tests whether the inductive bias embedded in our architecture (which assumes an underlying cluster structure) becomes detrimental when such structure is absent.

To evaluate the performance and scalability of our model, we consider four different problem scales. Each scale is characterized by the number of pickup--delivery pairs $n$ and the corresponding total number of customer nodes $2n$:
\[
n \in \{5, 10, 20, 40\} \quad \Longleftrightarrow \quad |\mathbf{P} \cup \mathbf{D}| \in \{10, 20, 40, 80\}.
\]
In the remainder of this section and in all tables, we refer to these problem classes as PDP10, PDP20, PDP40, and PDP80, where the number in the name indicates the total number of pickup and delivery nodes in the instance (i.e., PDP10 contains 5 pickup--delivery pairs, PDP80 contains 40 pairs). This notation is consistent across both clustered and uniform datasets (e.g., \texttt{pdp40-cluster}, \texttt{pdp80-uniform}).

Our policy network is based on the Transformer architecture described in Section~\ref{sec:methodology}. Unless otherwise stated, all hyper-parameters are kept fixed across different instance sizes and data distributions so that performance differences can be attributed primarily to the problem structure and the interaction with the learned policy. Specifically, the model's embedding dimension $d_h$ is set to $128$, the encoder contains $L=6$ dual-attention layers, and each multi-head attention module includes $M=8$ heads. In the final output layer of the decoder, we use a clipping parameter of $C=10$ (logit clipping) to stabilize the training process and to balance exploration and exploitation during decoding.

\begin{figure}[tbp]
    \centering
    \includegraphics[width=\columnwidth]{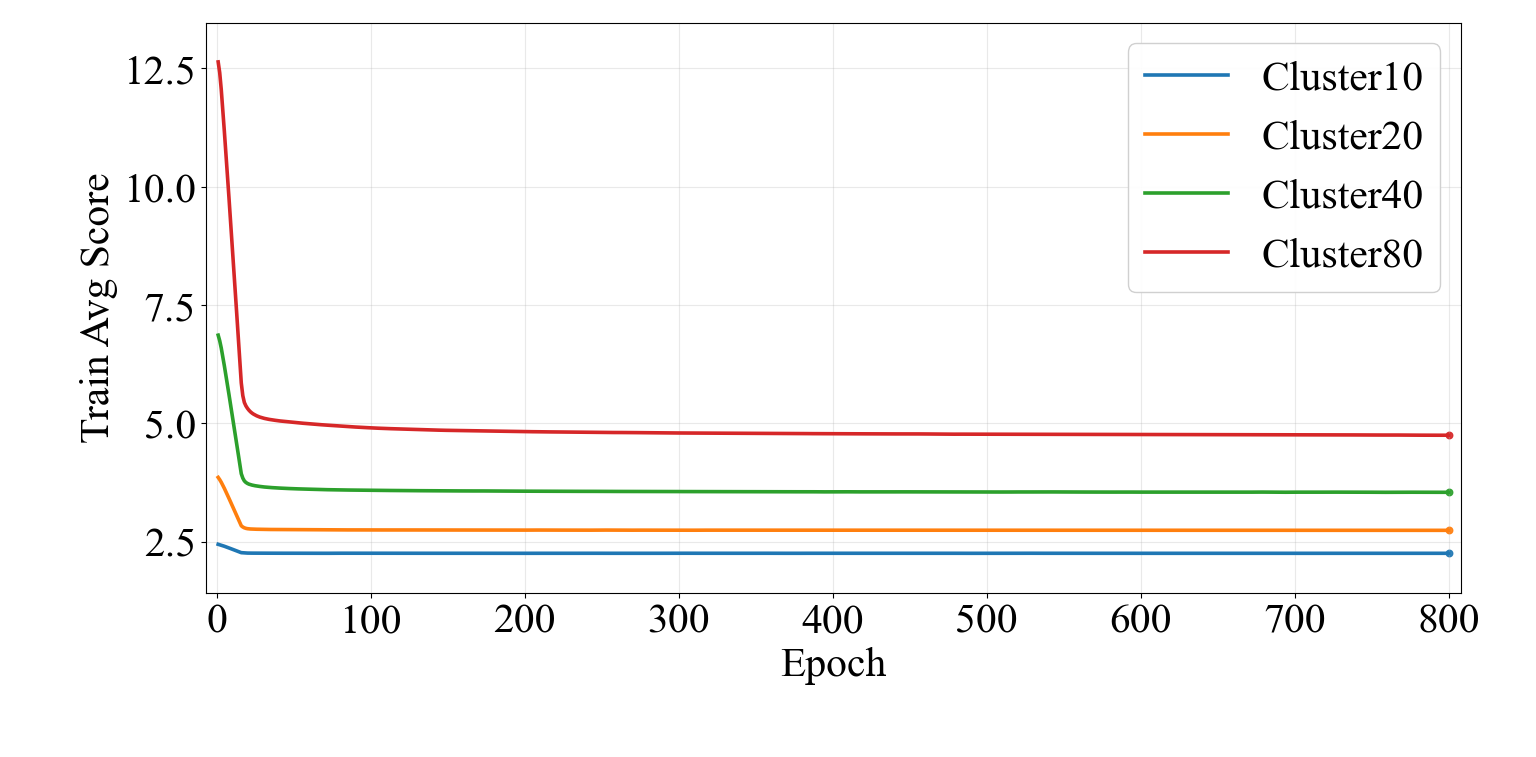}
    \caption{Training convergence on clustered instances.}
    \label{fig:train_convergence_cluster}
\end{figure}

During the training phase, problem instances are generated on the fly according to the target distribution (clustered or uniform). For each configuration, the model therefore encounters a large variety of instances differing in the relative positions of pickups, deliveries, and the depot, even when the problem size is fixed. We train for a total of $800$ epochs with a batch size of $512$. The model parameters are optimized using the Adam optimizer with a fixed learning rate of $1 \times 10^{-4}$. All implementations are based on the PyTorch framework. Training and test distributions are matched in terms of instance size and sampling scheme so that each reported result reflects out-of-sample performance on instances drawn from the same underlying distribution.

\subsection{Training Convergence and Stability}
\begin{figure}[tbp]
    \centering
    \includegraphics[width=\columnwidth]{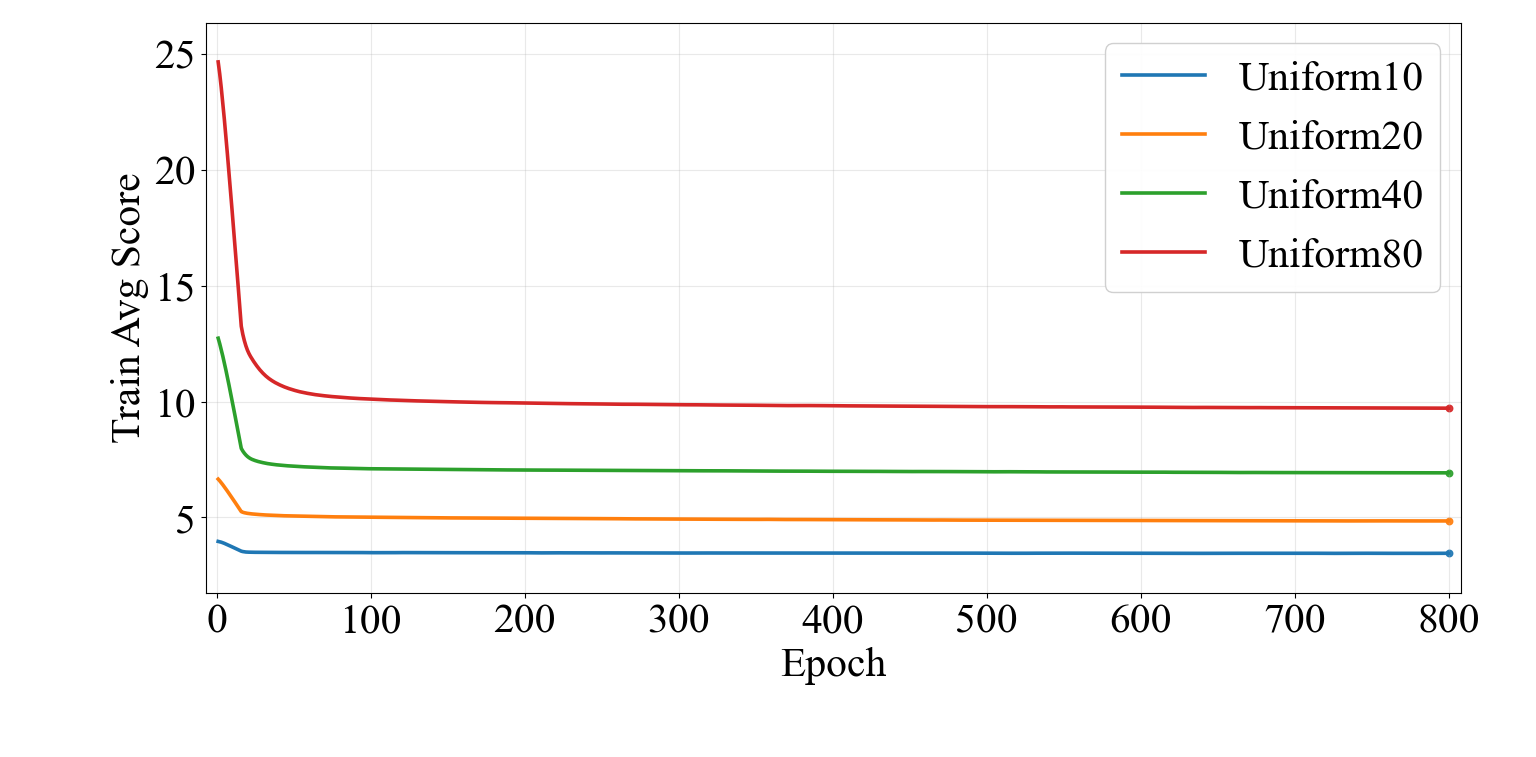}
    \caption{Training convergence on uniform instances.}
    \label{fig:train_convergence_uniform}
\end{figure}

In addition to final test performance, we monitor the training dynamics to verify optimization stability across distributions and problem scales. Figs.\ref{fig:train_convergence_cluster}--~\ref{fig:train_convergence_uniform} report the moving-average training objective over epochs for PDP10/20/40/80 under both uniform and clustered instance generators. Across all settings, the objective drops sharply in the early stage and then gradually saturates, indicating fast and stable convergence under the POMO-based policy gradient training. As expected, larger instances converge to higher objective values due to increased routing complexity, yet the learning curves remain smooth without oscillation or divergence. Notably, clustered instances consistently converge to lower objectives than uniform ones, reflecting that exploitable spatial structure makes the underlying routing task easier and aligns with the inductive bias of our cluster-aware encoder and hierarchical decoding.

%, 第一个表格：Cluster 数据集,
\begin{table*}[t]
\centering
\caption{Performance comparison on clustered PDP instances (lower objective is better).}
\label{tab:comparison_cluster}
{\setlength{\tabcolsep}{4pt}
\renewcommand{\arraystretch}{1.1}
\begin{adjustbox}{max width=\textwidth}
\begin{tabular}{
  @{}
  l
  S[table-format=2.3] c c
  S[table-format=2.3] c c
  S[table-format=2.3] c c
  S[table-format=2.3] c c
  @{}
}
\toprule
\multirow{2}{*}{\textbf{Method}} &
\multicolumn{3}{c}{\textbf{PDP10-cluster}} &
\multicolumn{3}{c}{\textbf{PDP20-cluster}} &
\multicolumn{3}{c}{\textbf{PDP40-cluster}} &
\multicolumn{3}{c}{\textbf{PDP80-cluster}} \\
\cmidrule(lr){2-4} \cmidrule(lr){5-7} \cmidrule(lr){8-10} \cmidrule(lr){11-13}
 & {Obj.} & {Gap (\%)} & {Time} & {Obj.} & {Gap (\%)} & {Time} & {Obj.} & {Gap (\%)} & {Time} & {Obj.} & {Gap (\%)} & {Time} \\
\midrule
CAADRL(greedy)      & 2.230 & 0.50 & \textbf{0.032} & 2.727 & 0.15 & \textbf{0.048} & 3.576 & 0.70 & 0.084 & 4.813 & 2.21 & 0.149 \\
CAADRL(Sample1280)  & 2.230 & 0.50 & 0.033 & 2.723 & \textbf{0.00} & 0.059 & 3.553 & 0.06 & 0.095 & 4.721 & 0.25 & 0.175 \\
CAADRL(Sample12800) & 2.230 & 0.50 & 0.033 & \textbf{2.723} & \textbf{0.00} & 0.056 & \textbf{3.551} & \textbf{0.00} & 0.098 & \textbf{4.709} & \textbf{0.00} & 0.198 \\
\midrule
ncs(t=1k)            & 2.224 & 0.23 & 0.073 & 2.733 & 0.37 & 0.081 & 3.694 & 4.03 & 0.097 & 4.802 & 1.97 & 0.164 \\
ncs(t=2k)            & \textbf{ 2.219} & \textbf{0.00} & 0.127 & 2.727 & 0.15 & 0.155 & 3.665 & 3.21 & 0.193 & 4.776 & 1.42 & 0.412 \\
ncs(t=3k)            & \textbf{ 2.219} & \textbf{0.00} & 0.202 & 2.724 & 0.04 & 0.233 & 3.649 & 2.76 & 0.287 & 4.734 & 0.53 & 0.444 \\
\midrule
Heter(greedy)        & 2.263 & 1.98 & 0.051 & 2.791 & 2.50 & 0.053 & 3.621 & 1.97 & \textbf{0.062} & 4.892 & 3.89 & \textbf{0.103} \\
Heter(Sample1280)    & 2.255 & 1.62 & 0.053 & 2.765 & 1.54 & 0.062 & 3.570 & 0.54 & 0.078 & 4.764 & 1.17 & 0.152 \\
Heter(Sample12800)   & 2.254 & 1.58 & 0.054 & 2.764 & 1.51 & 0.087 & 3.563 & 0.34 & 0.203 & 4.737 & 0.59 & 0.594 \\
\bottomrule
\end{tabular}
\end{adjustbox}}
\end{table*}

To validate the effectiveness of our model, we compare it with two existing state-of-the-art deep learning solvers for PDP:

\begin{itemize}
    \item \textbf{NCS} (\emph{ncs} in the tables): an efficient neural collaborative search framework that combines a neural construction policy with a neural neighborhood search procedure~\cite{kong2024ncs}. The parameter $t$ in \texttt{ncs(t=1k)}, \texttt{ncs(t=2k)}, and \texttt{ncs(t=3k)} denotes the number of improvement iterations performed during the collaborative search phase, with larger $t$ corresponding to more intensive local search.
    \item \textbf{Heter} (\emph{Heter} in the tables): a heterogeneous-attention encoder--decoder architecture that assigns different attention mechanisms to depot, pickup, and delivery nodes, thereby encoding precedence and pairing relationships at the attention level~\cite{li2022heter}.
\end{itemize}

These two methods are representative of, respectively, powerful neural improvement frameworks and specialized encoder--decoder constructions. Our model, denoted \textbf{CAADRL} in the tables, is evaluated under the same instance generation schemes and uses the same underlying training protocol (number of epochs, batch size, and optimizer settings) when applicable.

During the evaluation phase, we pre-generate a fixed test set of $100$ instances for each problem scale (PDP10, PDP20, PDP40, PDP80) and each data distribution (clustered and uniform). For each method and each configuration, we then report the average tour length over these $100$ instances and the average test time. All testing experiments were conducted independently on the vGPU-32GB card.

To comprehensively evaluate the impact of decoding strategies and sampling-based search, we employ three different decoding schemes for all neural construction models (CAADRL and Heter). For NCS, the reported scores correspond to the full collaborative search process as in its original design.

\textbf{Greedy.} At each decoding step, the model deterministically selects the node with the highest current probability. This strategy reveals the intrinsic quality of the learned policy without any additional search at inference time.

\textbf{Sampling-1280.} For each problem instance, we generate $1280$ solutions by performing $1280$ independent random samples from the probability distribution output by the policy network and return the best solution among them. This strategy allows the model to explore a richer subset of the solution space while keeping the computational cost moderate.

\textbf{Sampling-12800.} Similar to the previous strategy, but the number of samples is increased to $12800$, enabling a significantly larger search in the solution space. This configuration approximates the upper bound of what can be achieved by the learned policy under extensive stochastic sampling.

All reported results in the following subsections are the average tour lengths and average test time over the $100$ instances in the test set for each setting.

\begin{table*}[t]
\centering
\caption{Performance comparison on uniform PDP instances (lower objective is better).}
\label{tab:comparison_uniform}
{\setlength{\tabcolsep}{4pt}
\renewcommand{\arraystretch}{1.1}
\begin{adjustbox}{max width=\textwidth}
\begin{tabular}{
  @{}
  l
  S[table-format=2.3] c c
  S[table-format=2.3] c c
  S[table-format=2.3] c c
  S[table-format=2.3] c c
  @{}
}
\toprule
\multirow{2}{*}{\textbf{Method}} &
\multicolumn{3}{c}{\textbf{PDP10-uniform}} &
\multicolumn{3}{c}{\textbf{PDP20-uniform}} &
\multicolumn{3}{c}{\textbf{PDP40-uniform}} &
\multicolumn{3}{c}{\textbf{PDP80-uniform}} \\
\cmidrule(lr){2-4} \cmidrule(lr){5-7} \cmidrule(lr){8-10} \cmidrule(lr){11-13}
 & {Obj.} & {Gap (\%)} & {Time} & {Obj.} & {Gap (\%)} & {Time} & {Obj.} & {Gap (\%)} & {Time} & {Obj.} & {Gap (\%)} & {Time} \\
\midrule
CAADRL(greedy)      & 3.433 & 4.35 & \textbf{0.029} & 4.991 & 8.62 & \textbf{0.049} & 7.180 & 9.58 & 0.093 & 10.193 & 8.29 & 0.151 \\
CAADRL(Sample1280)  & 3.340 & 1.52 & 0.035 & 4.702 & 2.33 & 0.053 & 6.769 & 3.31 & 0.105 & 9.508  & 1.01 & 0.173 \\
CAADRL(Sample12800) & 3.330 & 1.22 & 0.042 & 4.661 & 1.44 & 0.067 & 6.702 & 2.29 & 0.109 & \textbf{9.413}  & \textbf{0.00} & 0.201 \\
\midrule
ncs(t=1k)            & 3.293 & 0.09 & 0.070 & 4.818 & 4.85 & 0.077 & 6.684 & 2.01 & 0.111 & 10.448 & 11.00 & 0.154 \\
ncs(t=2k)            & 3.292 & 0.06 & 0.141 & 4.804 & 4.55 & 0.157 & 6.589 & 0.56 & 0.215 & 10.205 & 8.41 & 0.306 \\
ncs(t=3k)            & \textbf{3.290} & \textbf{0.00} & 0.206 & 4.795 & 4.35 & 0.213 & \textbf{6.552} & \textbf{0.00} & 0.318 & 10.080 & 7.09 & 0.457 \\
\midrule
Heter(greedy)        & 3.380 & 2.74 & 0.055 & 4.890 & 6.42 & 0.081 & 7.228 & 10.32 & \textbf{0.074} & 10.547 & 12.05 & \textbf{0.108} \\
Heter(Sample1280)    & 3.303 & 0.40 & 0.063 & 4.623 & 0.61 & 0.069 & 6.802 & 3.82 & 0.079 & 10.166 & 8.00 & 0.149 \\
Heter(Sample12800)   & 3.297 & 0.21 & 0.065 & \textbf{4.595} & \textbf{0.00} & 0.092 & 6.849 & 4.53 & 0.202 & 10.101 & 7.31 & 0.596 \\
\bottomrule
\end{tabular}
\end{adjustbox}}
\end{table*}

\subsection{Comparison Analysis on Clustered Instances}

We first conduct a comprehensive performance comparison between our model and the baseline methods on the clustered datasets, which are the main focus of this work due to the strong alignment between our architectural design and the underlying data structure. All numerical results on clustered instances are summarized in Table~\ref{tab:comparison_cluster}.

From Table~\ref{tab:comparison_cluster}, several observations can be made:

\begin{itemize}
    \item \textbf{Small-scale instances (PDP10).} For the smallest clustered instances with 5 pickup--delivery pairs (10 customer nodes), NCS with intensive improvement iterations (\texttt{ncs(t=2k)} and \texttt{ncs(t=3k)}) slightly outperforms our model, achieving an average tour length of $2.219$ compared to $2.230$ for CAADRL. The relative gap is below $0.5\%$, indicating that on very small instances, where the search space is limited, the collaborative improvement mechanism of NCS can fully exploit local neighborhoods and close the gap to our specialized construction policy. Heter, on the other hand, lags behind both CAADRL and NCS at this scale, suggesting that role-specific attention alone is less effective than our explicit cluster-aware encoder.
    
    \item \textbf{Medium-scale instances (PDP20 and PDP40).} As the number of pairs increases to 10 (PDP20) and 20 (PDP40), our model becomes competitive or superior across the different decoding schemes. Under the largest sampling budget (Sampling-12800), CAADRL achieves the best scores for PDP20 and PDP40 (2.723 and 3.551, respectively), outperforming Heter (2.764 and 3.563) and NCS (2.724 and 3.649). The improvements over Heter are modest but consistent, particularly for PDP40, where CAADRL improves the average tour length by about $2.7\%$ relative to the best NCS configuration and slightly outperforms Heter. This confirms that the combination of global self-attention, intra-cluster attention, and hierarchical decoding provides a stronger inductive bias for medium-sized clustered PDP instances.

    \item \textbf{Large-scale instances (PDP80).} For the largest clustered instances with 40 pickup--delivery pairs, our model shows a relative advantage. Under Sampling-12800, the average tour length of CAADRL is 4.709, compared with 4.737 for Heter and 4.734 for NCS (\texttt{t=3k}). In addition, the average test time of our model on the clustered dataset is below 0.2 seconds, which is significantly shorter than that of the compared methods.
\end{itemize}

We also observe that, for both CAADRL and Heter, increasing the sampling scale from Greedy to Sampling-1280 and then to Sampling-12800 monotonically improves performance across all problem sizes. For example, on PDP80-cluster, our model improves from 4.806 (Greedy) to 4.721 (Sampling-1280) and finally to 4.709 (Sampling-12800), yielding a relative improvement of about $2\%$ compared to greedy decoding. This pattern confirms that the stochastic nature of the learned policy can be effectively exploited by simple sampling without any additional architectural changes, and that our model can provide multiple diverse high-quality solutions for the same instance. In addition, the average test time of our model is lower than that of the compared models in most settings.

Overall, the clustered experiments demonstrate that our model is particularly effective on exactly the kind of structured PDP instances it is designed for. The cluster-aware encoder and hierarchical dual-decoder with gating are able to capture the macro-level structure of the instance (two spatially separated clusters) and translate it into consistent intra-cluster routing and inter-cluster transitions. As the problem size increases, this inductive bias becomes increasingly beneficial, leading to clear gains over both role-specific attention (Heter) and neural collaborative search frameworks (NCS).

\begin{table*}[t]
\centering
\caption{Ablation study on clustered PDP instances (lower objective is better).}
\label{tab:ablation}
% 设置字体大小，\small 通常适合双栏表格，如果还觉得大可以用 \footnotesize
\small 
% 调整行高
\renewcommand{\arraystretch}{1.1}
% 设置列间距为0，完全由 extracolsep 自动填充
\setlength{\tabcolsep}{0pt} 

\begin{tabular*}{\textwidth}{
  @{\extracolsep{\fill}} % 关键：自动填充列间距
  l
  S[table-format=1.3] S[table-format=1.3] S[table-format=1.3]
  S[table-format=1.3] S[table-format=1.3] S[table-format=1.3]
  S[table-format=1.3] S[table-format=1.3] S[table-format=1.3]
  @{}
}
\toprule
\multirow{2}{*}{\textbf{Variant}} &
\multicolumn{3}{c}{\textbf{PDP20}} &
\multicolumn{3}{c}{\textbf{PDP50}} &
\multicolumn{3}{c}{\textbf{PDP100}} \\
\cmidrule(lr){2-4} \cmidrule(lr){5-7} \cmidrule(lr){8-10}
 & \multicolumn{1}{c}{Greedy} & \multicolumn{1}{c}{\makecell{Sample\\1280}} & \multicolumn{1}{c}{\makecell{Sample\\12800}} & \multicolumn{1}{c}{Greedy} & \multicolumn{1}{c}{\makecell{Sample\\1280}} & \multicolumn{1}{c}{\makecell{Sample\\12800}} & \multicolumn{1}{c}{Greedy} & \multicolumn{1}{c}{\makecell{Sample\\1280}} & \multicolumn{1}{c}{\makecell{Sample\\12800}} \\
\midrule
CAADRL     & \textbf{2.727} & \textbf{2.723} & \textbf{2.723} & \textbf{3.876} & \textbf{3.829} & \textbf{3.824} & \textbf{5.327} & \textbf{5.206} & 5.201 \\
no\_encoder & 2.763 & 2.759 & 2.758 & 3.907 & 3.861 & 3.857 & 5.341 & 5.236 & 5.220 \\
no\_decoder & 2.728 & 2.724 & \textbf{2.723} & 3.877 & 3.833 & 3.827 & 5.330 & 5.213 & \textbf{5.198} \\
POMO       & 2.731 & 2.727 & 2.726 & 3.882 & 3.832 & 3.826 & 5.407 & 5.262 & 5.236 \\
\bottomrule
\end{tabular*}
\end{table*}

\subsection{Generalization Analysis on Uniform Instances}
To evaluate the robustness of our cluster-aware design beyond its \emph{preferred} clustered setting, we conduct the same set of comparisons on uniformly generated PDP instances; the results are summarized in Table~\ref{tab:comparison_uniform}. Under the uniform distribution, pickup and delivery nodes are spatially mixed and no explicit region separation is present, which largely weakens the most direct inductive bias of our model (i.e., exploiting clear intra-/inter-cluster structure). Therefore, this setting provides a stringent test of whether the proposed architecture can still behave as a strong \emph{general-purpose} PDP solver rather than overfitting to clustered layouts.

Overall, CAADRL remains competitive across all sizes, and its relative advantages become more apparent as the problem scale increases.

\textbf{Small-scale uniform instances.}
For PDP10-uniform, NCS with intensive collaborative search (\,\texttt{ncs(t=3k)}\,) achieves the best objective value of $3.290$, while CAADRL reaches $3.330$ under Sampling-12800, corresponding to a small relative gap of $1.22\%$. Heter is also competitive at this scale ($3.297$ with Sampling-12800). This behavior is expected: when the instance is small, (i) the solution space is limited and NCS can effectively close the remaining gap via iterative improvements, and (ii) the benefit of explicitly separating intra-/inter-region decisions is less pronounced because long-range routing structure is not yet dominant.

\textbf{Medium-scale uniform instances.}
For PDP20-uniform and PDP40-uniform, the performance of all methods becomes closer and the best method depends on the scale. On PDP20-uniform, Heter obtains the best score ($4.595$ with Sampling-12800), and CAADRL achieves $4.661$ (gap $1.44\%$). On PDP40-uniform, NCS with \texttt{t=3k} is the strongest ($6.552$), while CAADRL attains $6.702$ (gap $2.29\%$). Importantly, although CAADRL is designed with cluster-aware components, it does not collapse on unstructured data: the gap to the best baseline remains within a few percent, suggesting that the global self-attention pathway and the dynamic fusion in the dual-decoder can still learn effective routing heuristics when explicit clusters disappear.

\textbf{Large-scale uniform instances.}
For PDP80-uniform, CAADRL becomes the best performer under Sampling-12800, achieving an average tour length of $9.413$, outperforming NCS (\texttt{t=3k}: $10.080$) and Heter (Sampling-12800: $10.101$). This corresponds to a relative improvement of about $6.6\%$ over the strongest baseline. This result indicates that the hierarchical design remains beneficial even in the uniform regime: as the number of requests grows, effective solutions increasingly exhibit \emph{multi-scale} structure (local exploitation plus occasional long-range transitions), and our architecture can capture such structure even without clearly separated geometric clusters.

\textbf{Effect of sampling and efficiency.}
Consistent with the clustered setting, increasing the sampling budget monotonically improves CAADRL on uniform instances. For example, on PDP80-uniform, the objective decreases from $10.193$ (Greedy) to $9.508$ (Sampling-1280) and further to $9.413$ (Sampling-12800), yielding a $7.7\%$ improvement over greedy decoding. Meanwhile, because CAADRL is a pure one-pass construction policy (without iterative neural improvement loops), its inference time under large sampling budgets is favorable compared with NCS, and remains practical even for PDP80.

Finally, comparing Tables~\ref{tab:comparison_cluster} and~\ref{tab:comparison_uniform}, all methods become worse on uniform instances, reflecting the increased difficulty without exploitable spatial regularities. Crucially, CAADRL exhibits a \emph{moderate} performance degradation and preserves competitive ranking across sizes, demonstrating that the cluster-aware encoder and hierarchical dual-decoder provide a useful inductive bias without sacrificing robustness on unstructured data.

\subsection{Ablation Study}
To validate the effectiveness of our architectural ingredients and disentangle the contributions of the encoder and decoder designs, we conduct an ablation study around the full model CAADRL. We consider three variants: \texttt{no\_encoder}, which replaces Cluster-Aware Attention with a standard Transformer encoder while keeping the Dynamic Dual-Decoder unchanged; \texttt{no\_decoder}, which removes the Dynamic Dual-Decoder and gating and uses a standard single decoder on top of the Cluster-Aware Encoder; and POMO, the vanilla POMO architecture with a standard encoder and a standard single decoder. All variants are evaluated on clustered test sets under the same decoding schemes as in Section~\ref{subsec:exp_settings} (Greedy, Sampling-1280, and Sampling-12800). In addition to the benchmark scales used above (PDP10/20/40/80), we further include larger instances PDP50 and PDP100 (i.e., 25 and 50 pickup--delivery pairs) to probe how the inductive biases embedded in the encoder and decoder scale with instance size. The results are summarized in Table~\ref{tab:ablation}.

Table~\ref{tab:ablation} demonstrates that the full model attains the best objective on PDP20 and PDP50 under all three decoding budgets, and also yields the strongest results on PDP100 under Greedy and Sampling-1280. Across all variants, increasing the sampling budget monotonically improves performance, and the absolute gain grows with the instance size; for example, CAADRL improves from 2.727 to 2.723 on PDP20 but from 5.327 to 5.206 on PDP100 when moving from Greedy to Sampling-1280, while the incremental improvement from Sampling-12800 is consistently smaller than the gain from Greedy to Sampling-1280. Under extensive sampling (Sampling-12800), \texttt{no\_decoder} becomes marginally better on PDP100 (5.198 vs.\ 5.201), indicating that best-of-$K$ selection can partly compensate for decoder design differences when the sampling budget is sufficiently large.

We first isolate the contribution of Cluster-Aware Attention in the encoder by comparing CAADRL against \texttt{no\_encoder}, where the Dynamic Dual-Decoder is kept unchanged. The degradation caused by removing Cluster-Aware Attention is consistent across all sizes and decoding schemes. Under Sampling-12800, the objective increases from 2.723 to 2.758 on PDP20 (about $+1.3\%$), from 3.824 to 3.857 on PDP50 (about $+0.9\%$), and from 5.201 to 5.220 on PDP100 (about $+0.4\%$). Since the decoder is identical in this comparison, these gaps can be cleanly attributed to the representational deficiency induced by replacing Cluster-Aware Attention with a standard Transformer encoder. This validates that explicitly decomposing interactions into global and intra-cluster components injects an effective inductive bias aligned with the clustered instance generator, leading to systematically better node embeddings for precedence-constrained routing decisions.

Next, we disentangle the impact of the Dynamic Dual-Decoder by comparing CAADRL with \texttt{no\_decoder}, which removes the dual-decoder and gating while keeping the Cluster-Aware Encoder intact. The resulting gains are modest on PDP20 and PDP50 (up to 0.004, around $0.1\%$), suggesting that once strong cluster-aware embeddings are available, a standard single decoder can already capture most routing regularities. However, the decoder contribution becomes more visible when inference is constrained to greedy or moderate sampling budgets on PDP100, where CAADRL improves over \texttt{no\_decoder} from 5.330 to 5.327 (Greedy) and from 5.213 to 5.206 (Sampling-1280). These improvements can be attributed to the gating mechanism that explicitly arbitrates between intra-cluster exploitation and inter-cluster transition decisions, reducing mode confusion in limited-search regimes. Under Sampling-12800, the difference reverses but remains within a very small margin (0.003), indicating diminishing marginal gains of decoder hierarchy once stochastic sampling provides sufficient exploration.

Finally, Table~\ref{tab:ablation} reveals an interaction between encoder- and decoder-side inductive biases. On PDP100, \texttt{no\_encoder} still outperforms the vanilla POMO baseline under sampling (e.g., 5.220 vs.\ 5.236 under Sampling-12800), which suggests that the Dynamic Dual-Decoder itself provides a complementary inductive bias at larger scales. Nevertheless, the best results consistently require the Cluster-Aware Encoder, and the full model offers the most robust performance profile across decoding budgets, demonstrating that the two components are largely complementary rather than redundant.

\begin{table*}[!t]
\centering
\renewcommand{\arraystretch}{1.25} % 调整行高，1.25 倍看起来最舒服
\setlength{\tabcolsep}{0pt} % 让表格自动撑满宽度

\caption{Cross-size generalization comparison on \textbf{Clustered} and \textbf{Uniform} instances (Lower objective is better).}
\label{tab:final_design}

\begin{tabular*}{\textwidth}{
  @{\extracolsep{\fill}}
  l 
  l 
  S[table-format=2.3] 
  S[table-format=2.3] 
  S[table-format=2.3] 
  S[table-format=2.3] 
  S[table-format=2.3] 
  S[table-format=2.3]
}
\toprule
\multirow{2}{*}{\textbf{Method}} & 
\multirow{2}{*}{\textbf{Decode Strategy}} & 
\multicolumn{3}{c}{\textbf{Clustered Instances}} & 
\multicolumn{3}{c}{\textbf{Uniform Instances}} \\
\cmidrule(lr){3-5} \cmidrule(l){6-8}
 & & {PDP200} & {PDP300} & {PDP500} & {PDP200} & {PDP300} & {PDP500} \\
\midrule

% Cluster 组
\multirow{3}{*}{CAADRL} 
 & Greedy          & 7.578 & {\textbf{9.601}} & {\textbf{13.205}} & 16.778 & 21.075 & 28.832 \\
 & Sampling-1280   & 7.507 & 9.834 & 14.811 & 15.974 & 21.099 & 31.297 \\
 & Sampling-12800  &  {\textbf{7.456}} & 9.754 & 14.653 & {\textbf{15.765}} & 20.798 & 30.917 \\

\midrule % 这里加上了你想要的横线分割

% POMO 组
\multirow{3}{*}{POMO} 
 & Greedy          & 7.798 & 10.116 & 14.484 & 16.958 & 21.465 & {\textbf{28.787}} \\
 & Sampling-1280   & 7.829 & 10.867 & 17.891 & 16.001 & 20.852 & 30.483 \\
 & Sampling-12800  & 7.750 & 10.736 & 17.630 & 15.816 & {\textbf{20.568}} & 30.058 \\

\bottomrule
\end{tabular*}
\end{table*}

\subsection{Cross-Size Generalization Across Problem Scales}
\begin{figure}[tbp]
    \centering
    \includegraphics[width=\columnwidth]{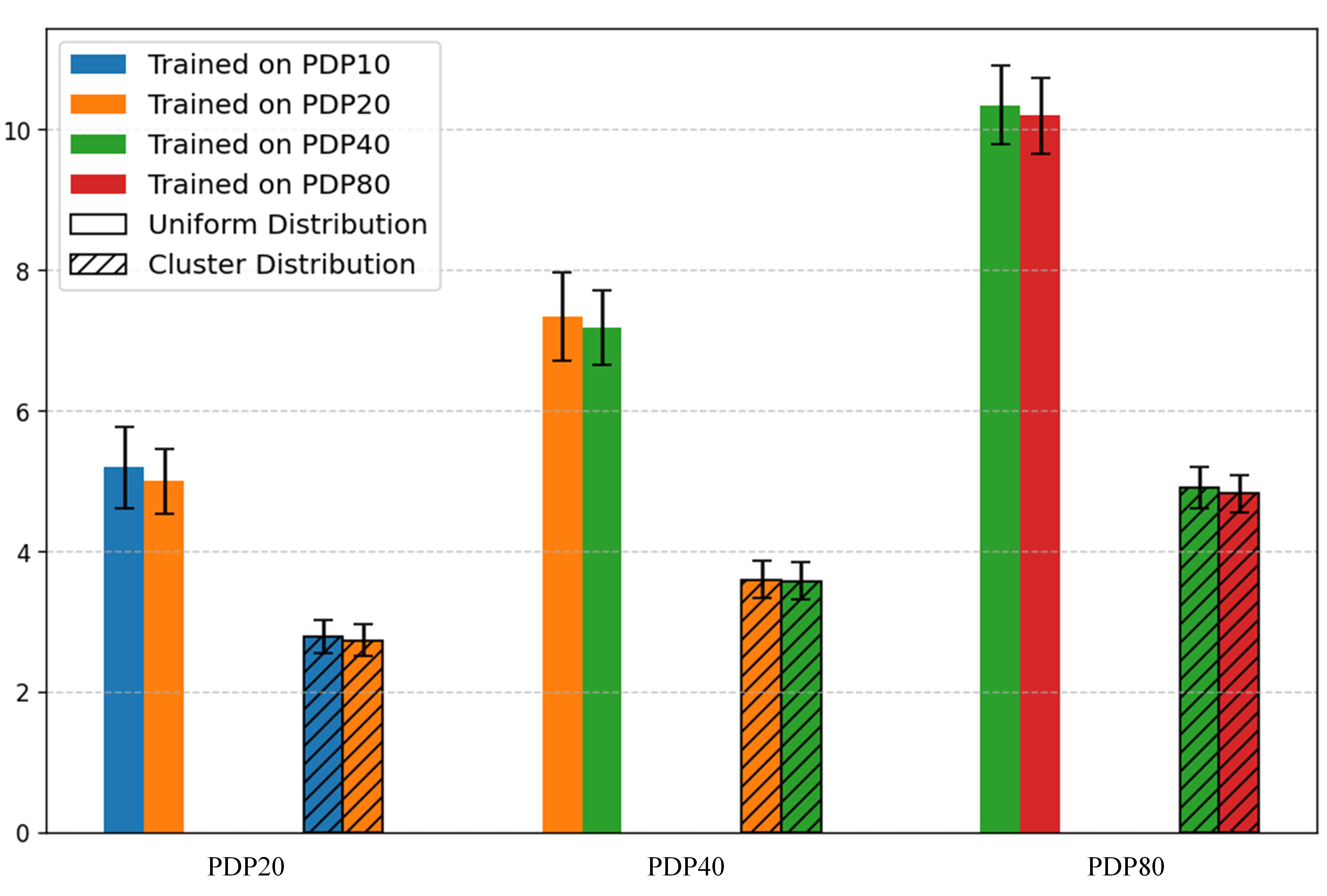}
    \caption{Cross-size generalization of \textbf{CAADRL} under greedy decoding.
    Solid bars indicate uniform instances and hatched bars indicate clustered instances. Colors denote the training problem size (PDP10/20/40/80). Error bars show the standard deviation over 100 test instances.}
    \label{fig:cross_size_generalization}
\end{figure}

Beyond distribution robustness, we further examine \emph{size generalization}: whether a policy trained on one problem scale can be directly transferred to larger instances \emph{without retraining}. Following the common practice in evaluating neural routing policies, we train CAADRL on PDP10/20/40/80 and evaluate it (with greedy decoding) on larger test sets PDP20/40/80 under both uniform and clustered distributions; the results are visualized in Fig.~\ref{fig:cross_size_generalization}.

Two clear trends emerge. First, the matched-size policy (training and testing on the same scale) consistently yields the best performance, confirming that the policy indeed adapts to the scale-dependent difficulty and tour structure. Second, the performance degradation under cross-size transfer is \emph{small and systematic} rather than catastrophic. On uniform instances, transferring PDP10 to PDP20, PDP20 to PDP40, and PDP40 to PDP80 increases the objective by about $4.0\%$, $2.2\%$, and $1.5\%$, respectively; on clustered instances, the corresponding increases are about $2.1\%$, $0.5\%$, and $1.7\%$. Notably, the transfer gaps tend to shrink for medium-to-large scales, suggesting that the learned representations capture routing principles that remain valid as the instance grows (e.g., balancing local continuity with global repositioning), which is precisely the kind of size robustness desired in practical deployment where request volumes can vary across time.

\subsection{Large-Scale Generalization from PDP100}
We further examine a more aggressive size transfer by training CAADRL on PDP100 and directly evaluating it on PDP200/300/500. POMO is reported as a generic reference baseline. Table~\ref{tab:final_design} summarizes results for both clustered and uniform distributions under the three decoding budgets.

\textbf{Clustered transfer.}
Table~\ref{tab:final_design} shows that CAADRL scales smoothly from PDP200 to PDP500 under all decoding budgets, without abrupt degradation. The advantage over the generic baseline widens with size, indicating that the cluster-aware encoder and hierarchical decoding learned at PDP100 capture scale-stable routing structure rather than size-specific heuristics. At PDP500, the greedy policy reduces the objective by about $8.8\%$ relative to POMO, and the sampling variants maintain a clear gap, which suggests that the learned intra-cluster and inter-cluster reasoning transfers to substantially larger clustered instances.

\textbf{Uniform transfer.}
On uniform instances (Table~\ref{tab:final_design}), CAADRL remains competitive across PDP200/300/500 and the greedy gaps stay within about $2\%$. This indicates that the cluster-aware inductive bias does not hinder extrapolation when explicit clusters disappear, and the policy trained at PDP100 provides stable large-scale performance.

\textbf{Sampling under large scale shift.}
Unlike the in-distribution setting, sampling does not consistently improve objectives when transferring from PDP100 to PDP300 or PDP500, and greedy decoding is often the strongest or close to the strongest option. This suggests that stochastic rollouts learned at PDP100 are less calibrated at much larger sizes, while the deterministic policy remains robust under extreme size extrapolation.

\section{Conclusion and Future Work}\label{sec:conclusion}

In this paper, we have proposed CAADRL, a deep reinforcement learning framework for the Pickup and Delivery Problem that is tailored to exploit clustered structure commonly observed in real-world instances. The central architectural contribution is a cluster-aware Transformer encoder that combines global self-attention with intra-cluster attention restricted by a cluster mask, thereby producing node embeddings that are simultaneously globally informative and locally role-aware. On top of these embeddings, we have introduced a hierarchical decoding mechanism with a Dynamic Dual-Decoder architecture and a learnable gating module. One decoder specializes in intra-cluster routing decisions, the other focuses on inter-cluster transitions, and the gating mechanism adaptively balances these two modes of behavior based on the current routing context. The entire policy is trained end-to-end under the POMO framework, which allows us to efficiently exploit solution symmetries and stabilize policy optimization by using multiple rollouts per instance as a shared baseline.

Extensive computational experiments on synthetically generated PDP testbeds have demonstrated the effectiveness of the proposed approach. On clustered instances, the primary scenario for which the architecture was designed, CAADRL achieves solution quality competitive with strong state-of-the-art baselines across a range of problem sizes, significantly outperforming a heterogeneous-attention encoder-decoder baseline and a powerful neural collaborative search method in medium and large-scale settings. In particular, our model shows clear advantages on instances with 40 pickup--delivery pairs, where the complexity of intra-cluster routing and inter-cluster transitions increases sharply, and the inductive bias towards cluster-awareness becomes especially beneficial. At the same time, on uniform instances without explicit cluster structure, the proposed method remains highly competitive: the performance gap to generic neural baselines is small for small and medium instances, and our model again takes the lead on the largest uniform instances considered. These results indicate that the additional modeling components introduced to exploit cluster structure do not compromise robustness, but instead provide a useful inductive bias that can still be leveraged when the data distribution is less structured.

Beyond the empirical performance gains, our study highlights a broader conceptual message: explicitly incorporating problem-specific structure, in this case, spatial clusters and the separation between intra-cluster and inter-cluster decisions into neural combinatorial optimization architectures can substantially enhance their scalability and generalization. Rather than relying solely on generic attention mechanisms and post-hoc feasibility masking, our design shows that carefully decomposing attention into global and intra-cluster components, and separating decision-making into local and global levels, can lead to more interpretable and more effective policies for complex routing problems such as PDP.

There are several promising directions for future work. First, while we have focused on a static, single-vehicle PDP with Euclidean distances, many real-world applications involve richer constraints. Synchronized transfers in microtransit planning couple the schedules of multiple vehicles through transfer timing, drone-assisted pickup--delivery couples ground routes with aerial sorties and coordination constraints between platforms, and multi-visit drone routing introduces repeated-service decisions driven by resource limits and trip coordination~\cite{fu2022microtransit,mulumba2024drone,meng2023multivisit}. These settings, together with multiple vehicles, time windows, heterogeneous fleets, and stochastic or dynamic request arrivals, provide natural testbeds for extending the proposed cluster-aware architecture and hierarchical decoding scheme, potentially by combining them with classical insertion heuristics or neural improvement modules. Second, in our current formulation, cluster information is implicitly defined by node roles (pickup vs.\ delivery) and spatial distributions. An interesting direction is to let the network learn latent cluster assignments directly from data, for example via differentiable clustering or graph partitioning layers, and to integrate these learned clusters into the attention and decoding process. Finally, it would be valuable to evaluate the proposed method on large-scale industrial datasets and to investigate system-level issues such as inference latency, deployment in rolling-horizon control frameworks, and interaction with human dispatchers or existing optimization pipelines.

Overall, the results in this paper suggest that cluster-aware attention and hierarchical decoding provide a powerful and flexible modeling paradigm for neural solvers of pickup-and-delivery-type routing problems. We believe that the ideas developed here can be fruitfully applied to a range of related combinatorial optimization tasks and can serve as a foundation for future research at the intersection of deep learning and large-scale transportation optimization.

\section*{Declarations}

\subsection*{Competing Interests}
The authors declare that they have no conflicts of interest.

\subsection*{Author Contributions}
Wentao Wang was responsible for drafting the manuscript and extending the original PyTorch source code to satisfy the pickup-and-delivery problem requirements. Lifeng Han contributed to the preparation of figures and was responsible for the implementation of baseline models and ablation studies. Guangyu Zou developed the original version of the reinforcement learning source code and supervised the entire study. Wentao Wang and Lifeng Han contributed equally to this work. All authors read and approved the final manuscript.

\subsection*{Ethical and Informed Consent for Data Used}
The data used in this study were generated randomly according to predefined rules. Therefore, no ethical approval or informed consent was required.

\subsection*{Data Availability and Access}
The data supporting the findings of this study are available from the corresponding author upon reasonable request.

\FloatBarrier
\nocite{ma2023neuopt}
\bibliography{references}

\end{document}